\documentclass[sigconf]{acmart}

\usepackage{booktabs}
\usepackage{multirow}
\usepackage{graphicx}  % for \resizebox
\usepackage{makecell} % in preamble

\AtBeginDocument{%
  }

\setcopyright{acmlicensed}
\copyrightyear{2025}
\acmYear{2025}
\acmDOI{XXXXXXX.XXXXXXX}
\acmConference[KDD 2025 Workshop]{3rd Workshop on Causal Inference and Machine Learning in Practice}{August 03--07,
  2025}{Toronto, CA}
\begin{document}

\title{Positive-Unlabeled Learning for Control Group Construction in Observational Causal Inference}

\author{Ilias Tsoumas}
% \authornote{Corresponding author}
\affiliation{%
  \institution{Artificial Intelligence Group, Wageningen University and Research}
  \country{The Netherlands}
}
\affiliation{%
  \institution{BEYOND Centre, National Observatory of Athens}
  \country{Greece}
}
\email{i.tsoumas@noa.gr}

\author{Dimitrios Bormpoudakis}
\affiliation{%
  \institution{BEYOND Centre, National Observatory of Athens}
  \country{Greece}
}

\author{Vasileios Sitokonstantinou}
\affiliation{%
  \institution{Image Processing Laboratory (IPL), University of Valencia}
  \country{Spain}
}

\author{Athanasios Askitopoulos}
\affiliation{%
  \institution{BEYOND Centre, National Observatory of Athens}
  \country{Greece}
}

\author{Andreas Kalogeras}
\affiliation{%
  \institution{BEYOND Centre, National Observatory of Athens}
  \country{Greece}
}

\author{Charalampos (Haris) Kontoes}
\affiliation{%
  \institution{BEYOND Centre, National Observatory of Athens}
  \country{Greece}
}

\author{Ioannis Athanasiadis}
\affiliation{%
  \institution{Artificial Intelligence Group, Wageningen University and Research}
  \country{The Netherlands}
}

\renewcommand{\shortauthors}{Tsoumas et al.}

\begin{abstract}
  In causal inference, whether through randomized controlled trials or observational studies, access to both treated and control units is essential for estimating the effect of a treatment on an outcome of interest. When treatment assignment is random, the average treatment effect (ATE) can be estimated directly by comparing outcomes between groups. In non-randomized settings, various techniques are employed to adjust for confounding and approximate the counterfactual scenario to recover an unbiased ATE. A common challenge, especially in observational studies, is the absence of units clearly labeled as controls—that is, units known not to have received the treatment. To address this, we propose positive-unlabeled (PU) learning as a framework for identifying, with high confidence, control units from a pool of unlabeled ones, using only the available treated (positive) units. We evaluate this approach using both simulated and real-world data. We construct a causal graph with diverse relationships and use it to generate synthetic data under various scenarios, assessing how reliably the method recovers control groups that allow estimates of true ATE. We also apply our approach to real-world data on optimal sowing and fertilizer treatments in sustainable agriculture. Our findings show that PU learning can successfully identify control (negative) units from unlabeled data based only on treated units and, through the resulting control group, estimate an ATE that closely approximates the true value.
  % We also discuss challenges related to the positivity assumption that arise in this context.
  This work has important implications for observational causal inference, especially in fields where randomized experiments are difficult or costly. In domains such as earth, environmental, and agricultural sciences, it enables a plethora of quasi-experiments by leveraging available earth observation and climate data, particularly when treated units are available but control units are lacking.
\end{abstract}






\maketitle

\section{Introduction}
% Soft intro - basic introduction to the field
% one line refer to the chanllenge/problem
% relevant \& related work that solve similar or the same issue
% PUATE + EDS with very similar way the same issue
% Contributions
% Stucture of the paper-sections
In modern science, when a randomized controlled trial is not a feasible option, we turn to observational studies to answer causal questions. The most common type of these questions is of the form  \textit{"What is the effect of intervention $T$ on outcome $Y$?"}. The average treatment effect (ATE) is a popular quantity that we estimate to answer these types of questions. Inferring causal ATE estimates of an intervention on any outcome formally requires access to control and treated units. Equally important for observational causal inference is the capability to adjust for other variables that are involved as confounders in the intervention-outcome system of interest, so that any difference in outcome between control and treated groups to be attributed only to the intervention.

% However, many issues arise, mainly in observational studies, related to data availability for unbiased ATE estimation. The most common issue is noisy or lacking observations for adjustment set covariates that are essential for the proper debiasing of ATE estimation. This set can be determined by direct covariate selection from domain experts or by the adjustment set that satisfies a graph-theoretic criterion (i.e. back-door criterion) applied on a causal graph. Matrix factorization and bayesian nonparametric generative models have been proposed as a solution to this lack or sparsity of observed covariates \cite{kallus2018causal, roy2018bayesian}.
% Another issue extends concerns that absence of data regarding either the treated or control groups.
The absence of treated or control group data is a critical concern. Molinari \cite{molinari2010missing} presents how we can realize estimation of ATE if we do not know the treatment status of a part of the units. Kuzmanovic et al. propose how to estimate conditional average treatment effects (CATE) \cite{kuzmanovic2023estimating} with missing treatment information. Beyond effect estimation under settings with partial absence of treatment information, Lancaster \& Imbens \cite{lancaster1996case}, building on \cite{steinberg1992estimating}, show that even when some control units are mistakenly classified and may have actually received the treatment—what they refer to as "contaminated controls"—it is still possible to recover unbiased estimates, as long as this misclassification is properly modeled in the analysis. Additionally, Rosenbaum \& Rubin exploit the availability of a pool of potential controls on building a balanced control group using multivariate matching methods that incorporate the propensity score, ensuring similarity between treated and control units based on observed covariates \cite{rosenbaum1985constructing}.

In this work, we focus on the case of the absolute absence of units that are clearly labeled as controls — that is, units known not to have received the treatment. To address this issue, we propose the use of positive-unlabeled (PU) learning \cite{bekker2020learning} as a formal framework for causal effect estimation in the absence of control units. Specifically, we identify, with high confidence, control units from a pool of unlabeled ones, using only the available treated (positive) units. Quite recently, PU learning has been proposed as a methodological solution to identify reliable control units from a pool of unlabeled one \cite{tsoumas2025leveraging}. In, parallel Kato et al. \cite{kato2025puate} propose a novel end-to-end PU learning method to estimate ATE in settings with a lack of control units. Complementary, with this work we especially contribute to this conceptual proposition through the following: i. Use PU learning to construct a reliable control group from unlabeled data de-linked from the effect estimation step. In this way, we facilitate the use of any causal model and adjustment set in the effect estimation part instead of the end-to-end solution of \cite{kato2025puate}. ii. We compare the use of an adjustment set defined by the back-door criterion for training the PU learner with a more exhaustive feature set that also includes \textit{bad controls}—such as mediators, colliders, and even the outcome variable itself, and show that the latter yields superior performance. iii. We perform experiments with simulated and real-world data around sustainable agriculture to showcase the usefulness of our approach.
    % \item Discussing and illustrating the possible issue with the hold of positivity assumption.
% \begin{itemize}
%     \item Use PU learning to construct a reliable control group from unlabeled data de-linked from the effect estimation step. In this way, we facilitate the use of any causal model and adjustment set in the effect estimation part instead of the end-to-end solution of \cite{kato2025puate}.
%     \item We compare the use of an adjustment set defined by the back-door criterion for training the PU learner with a more exhaustive feature set that also includes \textit{bad controls}—such as mediators, colliders, and even the outcome variable itself, and show that the latter yields superior performance.
%     \item We perform experiments with simulated and real-world data around sustainable agriculture to showcase the usefulness of our approach.
%     % \item Discussing and illustrating the possible issue with the hold of positivity assumption.
% \end{itemize}
These contributions aim to showcase the PU learning as a useful addition to the observational causal inference toolbox, specifically when real-world interventions have taken place, but we lack of labeled control units to realise an experiment. Especially, it applies in the domains of earth, environmental and agricultural sciences, where given the availability of earth observation and meteoclimatic data it can unlock a plethora of quasi-experiments, as we showcase in our real-world examples. 
\vspace{-0.3em}  % reduce space after the table
\section{Preliminaries \& Problem Formulation}

\begin{figure*}[t]
  \centering
  \includegraphics[width=\textwidth]{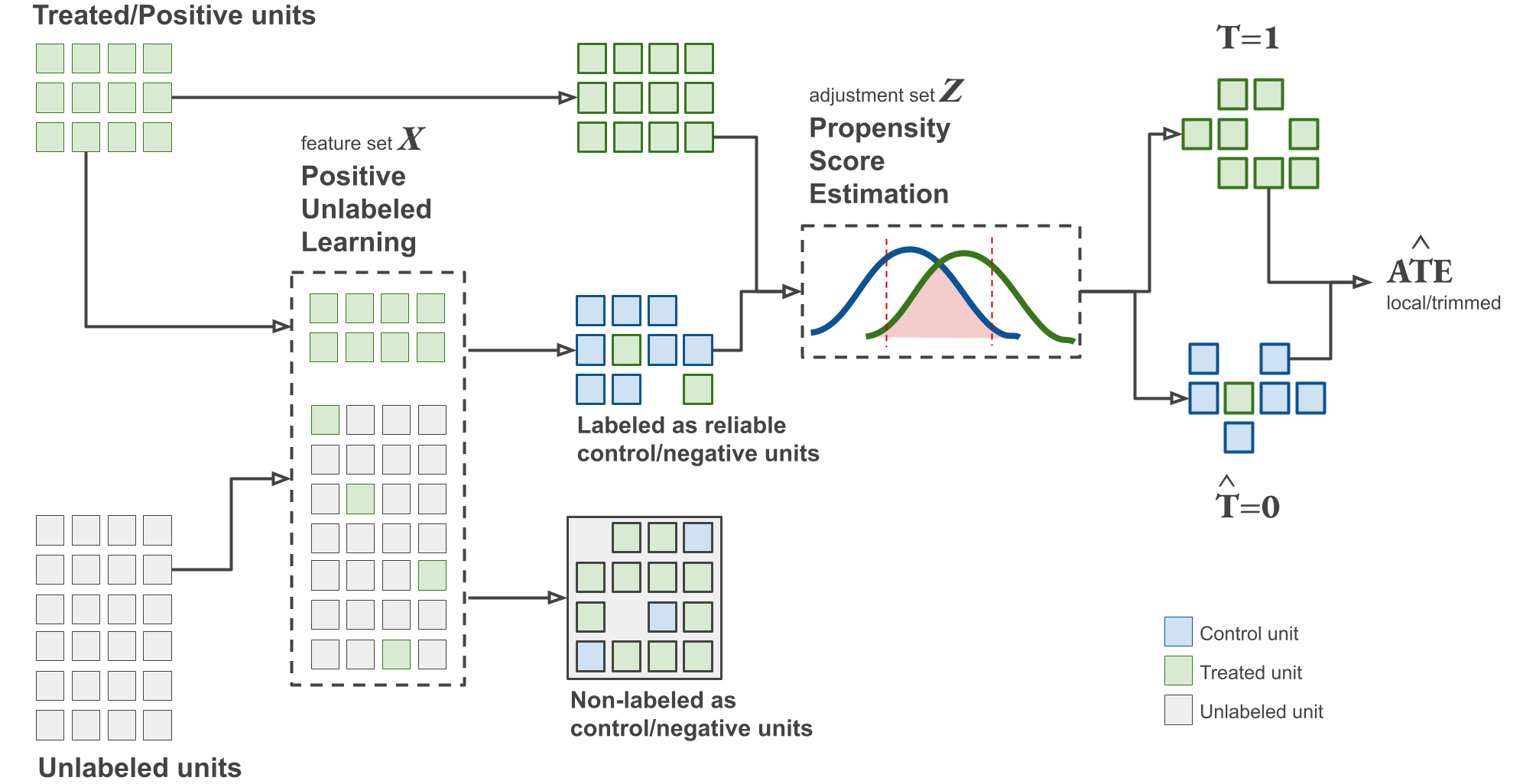}
  \caption{Overview of the use of PU learning as a preparatory step for causal estimation tasks that lack control groups.}
  \label{fig:pu-framework}
\end{figure*}

% Observational Causal Inference Assumptions \& our pipeline  :
% Graph creation - Effect identification - ATE Estimation.
% PU learning Assumptions \& our pipeline:
% Our case is Single-training-set scenario
% 2-step pu (spy + iterative svm) with adjustment vs with full feature set
\subsection{Observational Causal Inerence}
For each of the experimental setups that follow, simulated or real-world, we employ the relevant causal directed acyclic graph $G=(V,A)$ which includes all involved variables as vertices $V$  connected through directed edges $E$ and represent the causal relationships within the relevant treatment $T$ - outcome $Y$ system, e.g., the fertilizer - yield system. We limited all setups to binary treatments $T \in \{1,0\}$ and we aim to estimate their unbiased effects on an outcome of interest $Y$. We choose the exact adjustment set $Z \subseteq V$, if any, that satisfies the back-door criterion relative to (T, Y), blocking every path from $T$ to $Y$ that contains an arrow into $T$, and no descendant nodes from $Z$ to $T$ are allowed. Then, we retrieve $ATE$ as shown in Eq.~\ref{eq:ate-backdoor} based only on observational quantities.
\begin{align}
\text{ATE} &= \mathbb{E}[Y \mid do(T = 1)] - \mathbb{E}[Y \mid do(T = 0)] \nonumber \\
          &= \sum_z \left( \mathbb{E}[Y \mid T = 1, Z = z] - \mathbb{E}[Y \mid T = 0, Z = z] \right) \cdot P(Z = z)
\label{eq:ate-backdoor}
\end{align}
The capability to retrieve the $ATE$, whether we express it in Equation~\ref{eq:ate-backdoor} through as a structural causal model \cite{pearl2009causality}, or we express it via potential outcomes framework \cite{rubin2005causal}, rests on some fundamental assumptions that both frameworks directly or indirectly rely on: 
\begin{equation}
   \text{\textit{Unconfoundedness:} }Y(t) \Perp T \mid Z \quad \forall t \in\{0,1\}
   \label{eq:assum-unconfound}
\end{equation}
\begin{equation}
   \text{\textit{Positivity:} } 0<P(T=t \mid Z=z)<1 \quad \forall t \in\{0,1\}, \forall z \in Z
   \label{eq:assum-positivity}
\end{equation}
\begin{equation}
    \text{\textit{Consistency:} } Y=Y(t) \quad \text { if } T=t
    \label{eq:assum-consist}
\end{equation}

 Firstly, it is the stable unit treatment value assumption (SUTVA) that requires there are (i) no interference between units, i.e., one unit’s outcome is unaffected by other units’ treatment, and (ii) well-defined treatment with no different forms of each treatment level.  The unconfoundedness assumption (Eq.~\ref{eq:assum-unconfound}) requires all confounders are measured, and conditioning on them to removes bias. The positivity assumption (Eq.~\ref{eq:assum-positivity}) requires every unit has a positive probability of receiving each treatment level given adjustment set $Z$. The consistency assumption (Eq.~\ref{eq:assum-consist}) requires that when we observe $T=t$, the observed outcome $Y$ is equal to the outcome that would result from an intervention setting $T$ to $t$ through $d o(\cdot)$ operator. 

\subsection{PU learning for control group construction}
In this causal inference context, we should properly define the notation and slightly reformulate the assumptions of PU learning for the case of control group identification. In general, the typical target of PU learning is to learn a binary classifier using only a set of labeled positive instances and a set of unlabeled instances which include both positive and negative instances. In this work, we instrumentalize the PU learning framework as the means to recover from unlabeled units using only positive-treated units a subset of reliable negative-control units, for the downstream task of effect estimation.
% , and not for a binary classifier per se trained on firstly available postive and unlabeled data.

Thus, the target variable is the true class label, which in our case is the treatment $T \in \{1,0\}$ that indicates the real state of unit, where $T=1$ indicates a real treated/positive and $T=0$ indicates a real control/negative. $X$ represents the feature vector that describes all units and will be used to estimate $T$. A critical innovation we propose and explore in this work is that $X$ does not need to be limited to the aforementioned adjustment set $Z$; instead, in our formulation $X$ may contain any covariate included in $V$ (from causal graph $G$ that will be beneficial for a robust estimator of $T$) and even variables exogenous of $G$ system that will be usefull. So we state that $Z\subseteq V \subseteq X$. The variable that differentiates the notation from typical supervised learning is a third variable $S \in \{1,0\}$, which differs from $T$ that represents the real state of unit in terms of whether it has received or not treatment, $S$ is a label indicator. The  $S=1$ indicates that the unit is observed/annotated as treated/positive and $S=0$ indicates that unit is unlabaled, so it can be either a control/negative or a treated/positive.

\begin{equation}
    \text{\textit{SCAR:} }P(S=1 \mid T=1, X)=P(S=1 \mid T=1)=c \quad c \in (0,1)
    \label{eq:assum-pu-scar}
\end{equation}
\begin{equation}
    \text{\textit{No Label Noise in Treated:} }P(T=0 \mid S=1)=0
    \label{eq:assum-pu-onlytreatedlabeled}
\end{equation}
\begin{equation}
    \text{\textit{Controls in Unlabeled Set:} }P(T=0 \mid S=0)>0
    \label{eq:assum-pu-teleastonecontrol}
\end{equation}
\begin{equation}
\begin{aligned}
    \textit{Separability:} \\
    & P(t=1 \mid x)=\sigma(f(x)) \gg 0.5 \quad \text{for most } x \in \text{Treated} \\
                                 & P(t=0 \mid x)=\sigma(f(x)) \ll 0.5 \quad \text{for most } x \in \text{Controls}
\end{aligned}
\label{eq:assum-pu-seper}
\end{equation}
\begin{equation}
\begin{aligned}
    \textit{Smoothness:} \\
    &\left\|x-x^{\prime}\right\|<\varepsilon, \text{ for small } \epsilon>0 \\ 
    & \Rightarrow P(y=1 \mid x) \approx P\left(y=1 \mid x^{\prime}\right)
    \label{eq:assum-pu-smooth}
\end{aligned}
\end{equation}

The selected completely at random (SCAR) assumption (Eq.~\ref{eq:assum-pu-scar}) states that labeled positives are selected uniformly at random from all treated units, and selection for labeling is independent of features. The Eq.~\ref{eq:assum-pu-onlytreatedlabeled} implies that there is no possibility of mislabeling, e.g. a unit labeled as treated to be truly control. The Eq.~\ref{eq:assum-pu-teleastonecontrol} requires at least a control unit to be included in the unlabeled dataset. The separability assumption (Eq.~\ref{eq:assum-pu-seper}) state that there exists a decision function $f(x)$ that reliably separates the two classes among unlabeled units via score thresholding (ie $\sigma(\cdot)$ denotes the sigmoid function). The smoothness assumption (Eq.~\ref{eq:assum-pu-smooth}) that two units with nearby feature vectors $x$ $x'$ shared the same treatment status, which forces us to allow post-treatment and exogenous features to be included in $X$. Finally, the proposed end target of PU learner is to estimate the treatment assignment probability $P(T=1 \mid X)$ which allows to derive units that strongly looks truly untreated given its features (Eq.~\ref{eq:target-pu}) that can be used as controls in causal effect estimation.

\begin{equation}
    P(T=0 \mid X)=1-P(T=1 \mid X)\gg0.5
    \label{eq:target-pu}
\end{equation}

\section{Methodology \& Experimentation}
% Visualization of proposed framework
% Causal Graphs: simulated data, aaai, digestate (book of why) - 3 graphs
% back door criterion - adjustment set, 1st "real" ATE estimation with available data - 1 table

% PU with adjustment set vs full feature set(adjustment set + post-treatment,outcome)
% 1 table of them

% 4 propensity scores (spy set1, spy set2, svm set1, svm set2) plots

% triming - we should discuss the positivity assumption

% a BIG table with new ATEs per setup and the evaluation of pu learning where first we describe the evaluation process

% Maybe a plot that compare previous with after ATE.

% Finally the test with real unlabaled in digestate some evaluation metric, some visualization.
In this section, we describe in more detail: i. how we employ PU learning as a preparatory step to construct a reliable control group in the case of its absence to enable causal effect estimationl and ii. the four experimental setups - two simulated and two real-world datasets - where we test our idea.

\subsection{Framework for control group construction}
As Fig. ~\ref{fig:pu-framework} illustrates, we emphasize cases where treated and unlabeled units are available, in single-training-set scenarios either with simulated or real-world datasets, because we focus mainly in earth, environmental and agriculture cases where dominantly the units are pieces of land (e.g. agricultural parcels) that belong on the same population/dataset and either they receive an intervention/event or we do not know if they receive it or not. Specifically, we use a 2-step technique. In the first step we leverage the SPY method \cite{liu2002partially} where some of the real treated units are turned into spies, which we "hide" in the unlabeled dataset. A Naive Bayes classifier is trained, considering the unlabeled units as controls. Then, we label as "reliable" control units those for which the posterior probability is lower than the lowest of spies. In the second step, we employ iterative SVM (iSVM) \cite{yu2005single}. In each iteration, an SVM classifier is trained using the real treated units and the reliable controls from the first step. The unlabeled units that are classified as controls by this classifier are then added to the set of reliable controls for the next iteration till no class change occurs between two iterations.
Thus, we categorize the unlabeled units into two groups: the reliable control units, in which the PU learning assigned the control label with a tunable confidence; and the group with units that were left unlabeled because they do not differ enough from the real treated given features $X$. We do not consider the latter as treated because we already have a group with confirmed real treated units, so there is no reason to risk to add bias with some possible false labeled units as treated. Naturally, these two new groups will probably contain some false labeled units.
% - something that we need a way to evaluate. 

Afterwards, we train a propensity score estimator (i.e. a logistic regression) with the real treated units and the labeled from PU learning as reliable controls. However, we have already chosen the adjustment set $Z$ that satisfies the backdoor criterion based on the relevant causal graph $G$ that represents the system relationships for the estimation of the effect of $T$ on $Y$ and we train the propensity score estimator only using this subset of $X$, the adjustment set $Z$. We plot the propensity scores per group and we trim the scores of the two groups to ensure sufficient overlap between them. This trimming ensures that we avoid an extrapolation in regions with little comparability between groups, and we focus on $ATE$ estimation on the region of common support. Given that we have recovered as reliable controls the units that, intuitively speaking, significantly differ from the real treated for the PU learning estimator, which is trained on $X$ and in parallel we estimate propensity scores training an estimator on $Z$, it is clear that we risk drastic shrinkage of units in the area of overlap between the two groups due to the significant overlap of feature sets $Z\subseteq X$. This area of overlap is prone to be the area of the smallest propensity scores, given that reliable controls are units that are based on $X$ place in a conceptually similar area from the PU learning estimator. Consequently, an issue arises about how strong the positivity assumption holds each time and what exactly this strong \textit{'trimmed'} or \textit{'local'} $ATE$ quantity represents.
Finally, we estimate the $ATE$ with several estimators using the trimmed, based on propensity scores, treated and reliable control groups for reasons of completeness and framework assessment. We use linear regression and distance matching as baseline estimation methods, and Inverse Propensity Score Weigthing (IPW) \cite{stuart2010matching} and the causal machine learning method T-learner \cite{kunzel2019metalearners}. 

% \textbf{To Be Discussed
% }
% linear and non-linear setups with the bad \& good controls and graph for testing
% sowing setup
% digestate setup
\begin{figure}[!htbp]
  \centering
  \includegraphics[width=\columnwidth]{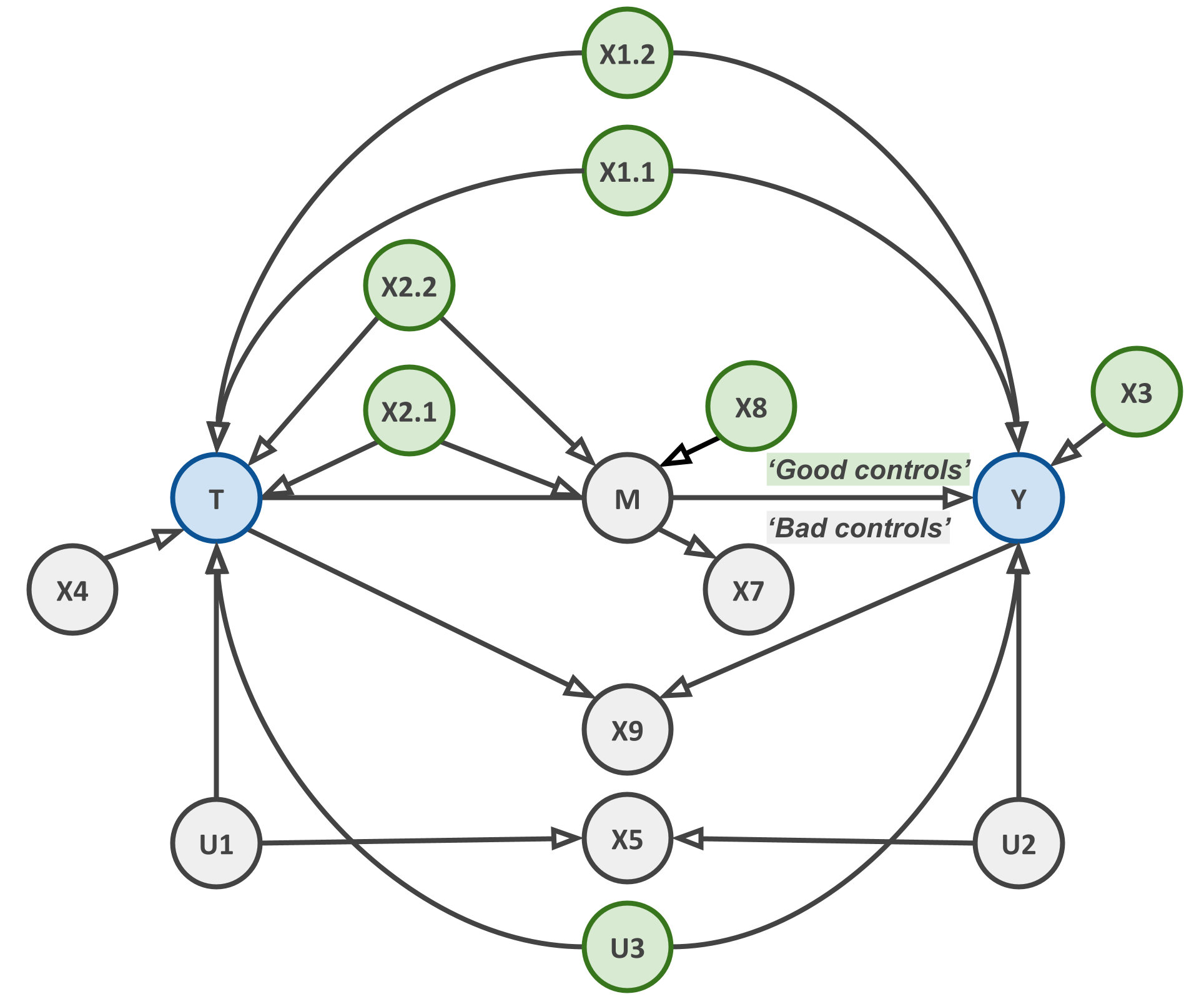}
  \caption{The causal graph $G_{sim}$ serves as the data-generating process for both linear and non-linear simulations.
  % It includes both good and bad controls to facilitate the exploration of proper set selection between $X,Z$.
  }
  \label{fig:sim-graph}
  \vspace{-1.5em}  % reduce space after the table
\end{figure}

% \usepackage{booktabs}
% \usepackage{multirow}
\begin{table*}[t]
\centering
\resizebox{\textwidth}{!}{%
\begin{tabular}{@{}ccccccccccccccc@{}}
\toprule
\multirow{3}{*}{\textbf{Dataset}} &
  \multirow{3}{*}{\textbf{PU method}} &
  \multirow{3}{*}{\textbf{Feature Set}} &
  \multirow{3}{*}{\textbf{\begin{tabular}[c]{@{}c@{}}\# Positives\\ (true treated)\end{tabular}}} &
  \multirow{3}{*}{\textbf{\# Spies}} &
  \multicolumn{2}{c}{\textbf{\# Unlabeled}} &
  \multicolumn{8}{c}{\textbf{Evaluation Metrics}} \\ \cmidrule(l){6-15} 
 &
   &
   &
   &
   &
  \multirow{2}{*}{\textbf{controls}} &
  \multirow{2}{*}{\textbf{treated}} &
  \multicolumn{2}{c}{\textbf{\# Selected}} &
  \multicolumn{2}{c}{\textbf{\# Non-selected}} &
  \multirow{2}{*}{\textbf{\begin{tabular}[c]{@{}c@{}}Control\\ Recall\end{tabular}}} &
  \multirow{2}{*}{\textbf{\begin{tabular}[c]{@{}c@{}}Control\\ Precision\end{tabular}}} &
  \multirow{2}{*}{\textbf{\begin{tabular}[c]{@{}c@{}}Contamination \\ Rate\end{tabular}}} &
  \multirow{2}{*}{\textbf{\begin{tabular}[c]{@{}c@{}}Treated\\ Leakage\end{tabular}}} \\ \cmidrule(lr){8-11}
 &
   &
   &
   &
   &
   &
   &
  \textbf{\# Controls} &
  \textbf{\# Treated} &
  \textbf{\# Controls} &
  \textbf{\# Treated} &
   &
   &
   &
   \\ \midrule
\multirow{4}{*}{\textbf{Linear}} &
  \multirow{2}{*}{\textbf{SPY}} &
  \textit{Z} &
  \multirow{4}{*}{346} &
  \multirow{4}{*}{103 (30\%)} &
  \multirow{4}{*}{505} &
  \multirow{4}{*}{149} &
  334 &
  31 &
  171 &
  118 &
  0.661 &
  0.915 &
  0.085 &
  0.208 \\
\textbf{} &
   &
  \textit{X} &
   &
   &
   &
   &
  505 &
  28 &
  0 &
  121 &
  1.000 &
  0.947 &
  0.053 &
  0.188 \\
\textbf{} &
  \multirow{2}{*}{\textbf{SPY+iSVM}} &
  \textit{Z} &
   &
   &
   &
   &
  143 &
  4 &
  362 &
  145 &
  0.283 &
  0.973 &
  0.027 &
  0.027 \\
\textbf{} &
   &
  \textit{X} &
   &
   &
   &
   &
  505 &
  0 &
  0 &
  149 &
  1.000 &
  1.000 &
  0 &
  0 \\ \midrule
\multirow{4}{*}{\textbf{Non-linear}} &
% \textbf{Non-linear} &
  \multirow{2}{*}{\textbf{SPY}} &
  \textit{Z} &
  \multirow{4}{*}{365} &
  \multirow{4}{*}{109 (30\%)} &
  \multirow{4}{*}{478} &
  \multirow{4}{*}{157} &
  186 &
  28 &
  292 &
  129 &
  0.389 &
  0.870 &
  0.131 &
  0.178 \\
\textbf{} &
   &
  \textit{X} &
   &
   &
   &
   &
  476 &
  15 &
  2 &
  142 &
  0.996 &
  0.969 &
  0.031 &
  0.095 \\
\textbf{} &
  \multirow{2}{*}{\textbf{SPY+iSVM}} &
  \textit{Z} &
   &
   &
   &
   &
  121 &
  16 &
  357 &
  141 &
  0.253 &
  0.883 &
  0.117 &
  0.102 \\
\textbf{} &
   &
  \textit{X} &
   &
   &
   &
   &
  474 &
  0 &
  4 &
  157 &
  0.992 &
  1.000 &
  0 &
  0 \\ \midrule
\multirow{4}{*}{\textbf{Sowing}} &
% \textbf{Sowing} &
  \multirow{2}{*}{\textbf{SPY}} &
  \textit{Z} &
  \multirow{4}{*}{35} &
  \multirow{4}{*}{5 (15\%)} &
  \multirow{4}{*}{121} &
  \multirow{4}{*}{15} &
  92 &
  3 &
  29 &
  12 &
  0.760 &
  0.968 &
  0.032 &
  0.200 \\
\textbf{} &
   &
  \textit{X} &
   &
   &
   &
   &
  72 &
  4 &
  49 &
  11 &
  0.595 &
  0.947 &
  0.053 &
  0.267 \\
\textbf{} &
  \multirow{2}{*}{\textbf{SPY+iSVM}} &
  \textit{Z} &
   &
   &
   &
   &
  97 &
  5 &
  24 &
  10 &
  0.802 &
  0.951 &
  0.049 &
  0.333 \\
\textbf{} &
   &
  \textit{X} &
   &
   &
   &
   &
  101 &
  5 &
  20 &
  10 &
  0.835 &
  0.953 &
  0.047 &
  0.333 \\ \midrule
\multirow{4}{*}{\textbf{Fertilization}} &
% \textbf{Fertilization} &
  \multirow{2}{*}{\textbf{SPY}} &
  \textit{Z} &
  \multirow{4}{*}{23} &
  \multirow{4}{*}{3 (15\%)} &
  \multirow{4}{*}{99} &
  \multirow{4}{*}{11} &
  64 &
  3 &
  35 &
  8 &
  0.646 &
  0.955 &
  0.045 &
  0.273 \\
\textbf{} &
   &
  \textit{X} &
   &
   &
   &
   &
  71 &
  5 &
  28 &
  6 &
  0.717 &
  0.934 &
  0.066 &
  0.455 \\
\textbf{} &
  \multirow{2}{*}{\textbf{SPY+iSVM}} &
  \textit{Z} &
   &
   &
   &
   &
  89 &
  7 &
  10 &
  4 &
  0.899 &
  0.927 &
  0.073 &
  0.636 \\
 &
   &
  \textit{X} &
   &
   &
   &
   &
  87 &
  8 &
  12 &
  3 &
  0.878 &
  0.916 &
  0.084 &
  0.727 \\ \bottomrule
\end{tabular}
}
\caption{Summary of control group selection using SPY and SPY+iSVM across four datasets (Linear, Non-linear, Sowing, Fertilization), evaluated with feature sets $Z$ (adjustment set) and $X$ (the most informative variables set about treatment).}
\label{tab:pu-evaluation-report}
\vspace{-2em}  % reduce space after the table
\end{table*}
\vspace{-0.8em}  % reduce space after the table

\subsection{Experimental setups \& data}
We simulate data under two different assumptions (generating $n = 1000$ samples for each), one with linear and one with non-linear causal relationships, to test our ideas in a fully controllable environment. For this purpose, we construct the causal graph $G_{sim}$ of Fig. \ref{fig:sim-graph} where we employ the Cinelli et al. work \cite{cinelli2024crash} in order to introduce $G_{sim}$ a wide pallette of different relationships \textit{'confounders', 'mediators', 'colliders'}. Specifically, vertices $V_{good}=\{X_{1.1}, X_{1.2}, X_{2.1}, X_{2.2}, U_{3}\}$ are \textit{'good controls'} reducing bias by blocking back-door paths if we control for them. The $V_{\text{bad}} = \{X_3,\allowbreak X_5,\allowbreak X_7,\allowbreak X_9,\allowbreak M\}$ are \textit{'bad controls'}, with addition of $X_3$ as control would lead to \textit{'bias amplification'}. The $X_{5}$ are known as \textit{'M-bias'} that induce bias through $U_1, U_2$. $M$ mediates the effect we want to estimate, so we keep this path untouched without controlling for this variable, similarly to $X7$, where a control for it is equivalent to a partial control for mediator $M$. $X9$ is a typical \textit{'bad control'} because controlling for it opens a colliding path and induces \textit{'selection bias'}. Finally, vertices $V_{neutral}=\{X_{3},  X_{8}\}$ are neutral in terms of inserting or removing bias but can be useful controls in terms of $ATE$ precision. Thus, as expected from graph construction, the minimum adjustment set $Z_{sim}$ that satisfies the back-door criterion contains the \textit{'good controls'} so $Z_{sim} = V_{good} =\{X_{1.1}, X_{1.2}, X_{2.1}, X_{2.2}, U_{3}\}$ for an unbiased estimation of $ATE$ of treatment $T$ on outcome $Y$ of causal system $G_{sim}$. The Sec.~\ref{sec:sim-setup} of the Appendix present the data-generating process in detail. 

Also, we investigate the applicability of our ideas in two real-world scenarios from the domain of sustainable agriculture. First, we use the causal graph, data and and causal effect estimation setup from Tsoumas et al. work \cite{tsoumas2023evaluating}. They estimate the $ATE$ of treatment $T$: whether the cotton field is sown on an annotated date as favorable or not, on the outcome $Y$, which is the final yield of cotton. We use the adjustment set $Z_{sowing}= $ \textit{\{Weather on sowing date, Soil moisture on sowing date, Topsoil properties, Topsoil organic carbon, Seed variety, Geometry of field} that satisfies the back-door criterion of their proposed causal graph, and also we use the additional available variables $V_{extra}=$ \textit{\{Crop growth, Location \& area of field, Harvest date \& cultivation period, Yield\}} that are not used for effect estimation in order to test the use of an expanded feature set $X_{sowing}=Z_{sowing}+V_{extra}$ over the adjustment set $Z_{sowing}$ to construct a group of reliable control units and the hold of the positivity assumption. As a second real-world scenario, we introduce a new dataset about the impact of a digestate fertilizer (i.e., a sustainable alternative to chemical fertilizers) on wheat biomass in fields cultivated with durum wheat. For this paper, we adopt the popular example of fertilization-yield causal graph from Pearl's book \cite{pearl2018book}, incorporate it with the richer ground truth causal graph in \cite{sharma2023knowledge} regarding soil processes. In this dataset, the adjustment set and the feature set are composed as follows: $Z_{fertilizer}= $\textit{\{Topsoil organic carbon, Accumulated\ precipitation, Growing degree days, Accumulated vegetation moisture content, Soil moisture, Soil type Seed variety\}} and $X_{sowing}=Z_{sowing}+$ \textit{\{Exogenous organic matter indexes, Produced biomass proxy\}}. More details about the effect estimation and the data in the Table~\ref{tab:rworld-data-vars} on Sec.~\ref{sec:rworld-data} of Appendix.
\vspace{-0.85em}  % reduce space after the table

\section{Results}
\begin{table*}[!htbp]
\centering
\resizebox{\textwidth}{!}{%
\begin{tabular}{@{}lllllllllllllllll@{}}
\toprule
\multicolumn{1}{c}{\multirow{3}{*}{\textbf{Dataset}}} &
  \multicolumn{1}{c}{\multirow{3}{*}{\textbf{Control Units}}} &
  \multirow{3}{*}{\textbf{(treated, control)}} &
  \multirow{3}{*}{\textbf{true effect}} &
  \multirow{3}{*}{\textbf{PU Feature Set}} &
  \multicolumn{12}{c}{\textbf{Causal Effect Estimation Methods}} \\ \cmidrule(l){6-17} 
\multicolumn{1}{c}{} &
  \multicolumn{1}{c}{} &
   &
   &
   &
  \multicolumn{3}{c}{\textbf{Linear Regression}} &
  \multicolumn{3}{c}{\textbf{IPS weighting}} &
  \multicolumn{3}{c}{\textbf{Matching}} &
  \multicolumn{3}{c}{\textbf{T-Learner(RF)}} \\ \cmidrule(l){6-17} 
\multicolumn{1}{c}{} &
  \multicolumn{1}{c}{} &
   &
   &
   &
  \textbf{ATE} & \textbf{CI} & \textbf{p-value} &
  \textbf{ATE} & \textbf{CI} & \textbf{p-value} &
  \textbf{ATE} & \textbf{CI} & \textbf{p-value} &
  \textbf{ATE} & \textbf{CI} & \textbf{p-value} \\ \midrule
\multirow{5}{*}{\textbf{Linear}} &
  \textbf{real controls} &
  (495, 505) - no trim &
  \multirow{5}{*}{3.000} &
  - &
  3.664 &
  (3.447, 3.895) &
  0.000 &
  3.848 &
  (3.451, 4.237) &
  0.001 &
  3.927 &
  (3.601, 4.192) &
  0.001 &
  4.029 &
  (3.939, 4.119) &
  - \\
 &
  \multirow{2}{*}{\textbf{SPY}} &
  (36, 232) &
   &
  \textit{Z} &
  2.718 &
  (2.250, 3.143) &
  0.000 &
  3.081 &
  (1.927, 4.121) &
  0.001 &
  3.261 &
  (2.648, 3.620) &
  0.001 &
  3.520 &
  (3.211, 3.828) &
  - \\
 &
   &
  (450, 414) &
   &
  \textit{X} &
  2.800 &
  (2.631, 2.997) &
  0.000 &
  2.818 &
  (2.337, 3.376) &
  0.001 &
  3.282 &
  (2.989, 3.392) &
  0.001 &
  3.529 &
  (3.428, 3.630) &
  - \\
 &
  \multirow{2}{*}{\textbf{SPY+iSVM}} &
  (22, 147) &
   &
  \textit{Z} &
  2.757 &
  (2.109, 3.338) &
  0.000 &
  3.231 &
  (1.817, 4.500) &
  0.001 &
  3.667 &
  (2.906, 4.250) &
  0.001 &
  3.771 &
  (3.331, 4.212) &
  - \\
 &
   &
  (414, 424) &
   &
  \textit{X} &
  3.021 &
  (2.861, 3.185) &
  0.000 &
  3.188 &
  (2.765, 3.559) &
  0.001 &
  3.574 &
  (3.320, 3.726) &
  0.001 &
  3.814 &
  (3.720, 3.909) &
  - \\ \midrule
\multirow{5}{*}{\textbf{Non-linear}} &
  \textbf{real controls} &
  (522, 478) - no trim &
  \multirow{5}{*}{9.525} &
  - &
  9.717 &
  (9.367, 10.094) &
  0.000 &
  9.799 &
  (9.296, 10.229) &
  0.001 &
  9.597 &
  (9.155, 9.869) &
  0.001 &
  9.829 &
  (9.646, 10.011) &
  - \\
 &
  \multirow{2}{*}{\textbf{SPY}} &
  (17,77) &
   &
  \textit{Z} &
  8.057 &
  (5.989, 10.452) &
  0.000 &
  7.775 &
  (4.819, 10.827) &
  0.001 &
  7.209 &
  (4.660, 8.793) &
  0.001 &
  7.261 &
  (6.322, 8.201) &
  - \\
 &
   &
  (513, 487) &
   &
  \textit{X} &
  9.480 &
  (9.100, 9.886) &
  0.000 &
  9.570 &
  (9.040, 10.035) &
  0.001 &
  9.321 &
  (8.900, 9.607) &
  0.001 &
  9.324 &
  (9.132, 9.516) &
  - \\
 &
  \multirow{2}{*}{\textbf{SPY+iSVM}} &
  (86, 137) &
   &
  \textit{Z} &
  7.962 &
  (6.762, 8.993) &
  0.000 &
  7.884 &
  (6.269, 9.466) &
  0.001 &
  7.556 &
  (6.318, 8.178) &
  0.001 &
  7.877 &
  (7.384, 8.371) &
  - \\
 &
   &
  (513, 470) &
   &
  \textit{X} &
  9.719 &
  (9.378, 10.0953) &
  0.000 &
  9.820 &
  (9.369, 10.254) &
  0.001 &
  9.805 &
  (9.446, 10.092) &
  0.001 &
  9.887 &
  (9.701, 10.073) &
  - \\ \midrule
\multirow{5}{*}{\textbf{Sowing}} &
  \textbf{real controls} &
  (50, 121) &
  \multirow{5}{*}{\makecell{expected \\ positive}} &
  - &
  546 &
  (211, 880) &
  0.002 &
  471 &
  (138, 816) &
  0.001 &
  448 &
  (186, 760) &
  0.006 &
  372 &
  (215, 528) &
  0.024 \\
 &
  \multirow{2}{*}{\textbf{SPY}} &
  (13, 92) &
   &
  \textit{Z} &
  211 &
  (-479, 902) &
  0.544 &
  315 &
  (-487, 1024) &
  0.138 &
  244 &
  (-733, 1099) &
  0.226 &
  259 &
  (-134, 653) &
  - \\
 &
   &
  (14, 14) &
   &
  \textit{X} &
  664 &
  (-267, 1596) &
  0.147 &
  811 &
  (122, 1660) &
  0.030 &
  689 &
  (-54, 1369) &
  0.063 &
  777 &
  (376, 1178) &
  - \\
 &
  \multirow{2}{*}{\textbf{SPY+iSVM}} &
  (25, 43) &
   &
  \textit{Z} &
  156 &
  (-463, 776) &
  0.615 &
  322 &
  (-25, 724) &
  0.087 &
  293 &
  (-145, 723) &
  0.128 &
  306 &
  (82, 530) &
  - \\
 &
   &
  (39, 47) &
   &
  \textit{X} &
  466 &
  (-61, 993) &
  0.082 &
  524 &
  (170, 992) &
  0.029 &
  480 &
  (132, 857) &
  0.015 &
  390 &
  (195, 586) &
  - \\ \midrule
\multirow{5}{*}{\textbf{Ferilization}} &
  \textbf{real controls} &
  (34, 99) - no trim &
  \multirow{5}{*}{\makecell{expected \\ positive}} &
  - &
  1.188 &
  (-0.386, 2.762) &
  0.138 &
  1.160 &
  (-0.094, 2.649) &
  0.055 &
  0.708 &
  (-0.834, 2.266) &
  0.200 &
  1.104 &
  (0.484, 1.725) &
  - \\
 &
  \multirow{2}{*}{\textbf{SPY}} &
  (8, 38) &
   &
  \textit{Z} &
  3.000 &
  (-0.600  6.599) &
  0.097 &
  1.866 &
  (-0.173, 4.351) &
  0.094 &
  1.111 &
  (-1.188, 2.842) &
  0.219 &
  0.903 &
  (-0.141, 1.947) &
  - \\
 &
   &
  (20, 47) &
   &
  \textit{X} &
  2.342 &
  (-0.168, 4.851) &
  0.067 &
  0.805 &
  (-0.711, 2.604) &
  0.167 &
  0.425 &
  (-1.173, 2.268) &
  0.345 &
  0.464 &
  (-0.363, 1.292) &
  - \\
 &
  \multirow{2}{*}{\textbf{SPY+iSVM}} &
  (24, 67) &
   &
  \textit{Z} &
  1.302 &
  (-0.566, 3.170) &
  0.168 &
  1.190 &
  (-0.391, 2.843) &
  0.076 &
  0.981 &
  (-0.426, 3.142) &
  0.162 &
  0.542 &
  (-0.228, 1.312) &
  - \\
 &
   &
  (31, 67) &
   &
  \textit{X} &
  1.750 &
  (0.003, 3.498) &
  0.049 &
  1.322 &
  (-0.347, 3.122) &
  0.047 &
  0.742 &
  (-0.555, 2.265) &
  0.192 &
  0.823 &
  (0.158, 1.488) &
  - \\
\bottomrule
\end{tabular}%
}
\caption{
Effect estimation using treated units and (i) real control, (ii) retrieved reliable controls under various PU learning methods, feature sets, and ATE estimation techniques.
% The effect estimation with i. the real control and treated units \& ii. retrieved reliable controls that has considered as unlabaled alongside with a 30\% portion of real treated under various variatons on PU learning method, feature set for PU learning and estimation method of ATE
}
\label{tab:pu-ate-before-after}
\vspace{-2em}  % reduce space after the table
\end{table*}

% \begin{figure}[t]
%   \centering
%   \includegraphics[width=\columnwidth]{combined_propensity_scores_2x2.png}
%   \caption{Propensity score overlap in the non-linear setup varies by predictor set and method, affecting positivity and common support.}
%   \label{fig:pu-ps_after_pu}
% \end{figure}

% Interpretation of results - with framework's steps order

% Evaluation metric only how good PU learning retrieve real controls - with hide\& seek reasoning and in real situation where no controls. ('control recall','control purity','contamination rate','treated leakage')
% \ref{tab:pu-evaluation-report}

% 4 propensity scores (spy set1, spy set2, svm set1, svm set2) plots
% triming - we should discuss the positivity assumption
% \ref{fig:pu-ps_after_pu}

% alongside the graph of explainability.

% PU with adjustment set vs full feature set(adjustment set + post-treatment,outcome)
% \ref{tab:pu-ate-before-after}

% 4 propensity scores (spy set1, spy set2, svm set1, svm set2) plots
% triming - we should discuss the positivity assumption
% \ref{fig:pu-ps_after_pu}

% alongside the graph of explainability.

% Maybe a plot that compare previous with after ATE.

% Finally the test with real unlabaled in fertilization case some evaluation metric

% Maybe some map with spatial visualization about constructed control for aaai and digestate. Maybe only in the real unlabaled case.
In this section we present \& discuss the results of our experiments regarding the proposed utilization of PU learning as an enabler of any causal effect estimation task when a confirmed control group is completely absent, but a source of unlabeled units is available.
% but a source of possible control units is available as unlabeled units.

For the evaluation of the framework, we emulate PU conditions through the popular \textit{'hide and seek'} approach \cite{saunders2022evaluating}. Specifically, we engineer positive-unlabeled datasets from the aforementioned simulated and real-world positive-negative datasets. We do this simply by hiding a percentage of positive/treated units within the negative/control units, something that allows us to consider this mixed group with units from both treatment classes as an unlabelled set to test our ideas. This satisfies the SCAR assumption (Eq. \ref{eq:assum-pu-scar}) and facilitates the assessment of the PU learner using common binary classification evaluation metrics. However, we have to clarify some slight conceptual changes from typical classification metrics that better align with and reflect a causal inference context. While throughout the paper we refer to treated units as positives and controls as negatives in the PU learning setup, during evaluation, we treat true controls as the positive class to assess how well the method recovers clean / reliable control units. Thus, for clarity, under this evaluation perspective, True Positives (TP) correspond to true control units correctly recovered as controls, False Positives (FP) correspond to treated units wrongly selected as controls, False Negatives (FN) to true control units that were not recovered and True Negatives (TN) to treated units correctly not selected (held out as no control units). Thus, we use: the Control Recall equation (Eq. \ref{eq:eval-recall}) to assess how many of the true controls were recovered (it measures the coverage of the control group);  the Control Precision equation (Eq. \ref{eq:eval-precision}), which summarizes how many were actual controls out of selected as 'reliable controls'; the Contamination Rate (Eq. \ref{eq:eval-contam}), which refers to the proportion of selected as 'reliable controls' that were actually treated; and we introduce the Treated Leakage metrics (Eq. \ref{eq:eval-leakage}) that measures how many units mistakenly ended up in 'reliable controls' among true treated units.

\begin{equation}
    \text { Control Recall } = \frac{T P}{T P+F N}
    \label{eq:eval-recall}
\end{equation}
\begin{equation}
    \text { Control Precision } = \frac{T P}{T P+F P}
    \label{eq:eval-precision}
\end{equation}
\begin{equation}
    \text { Contamination Rate } = \frac{F P}{T P+F P}
    \label{eq:eval-contam}
\end{equation}
\begin{equation}
    \text { Treated Leakage } = \frac{F P}{F P+T N}
    \label{eq:eval-leakage}
\end{equation}

In Table \ref{tab:pu-evaluation-report}, the evaluation metrics of PU learning for retrieving real control units from an unlabeled pool of mixed control and treated units are presented in detail for the four different datasets. In the two simulated, linear and non-linear, datasets our results show that the combination of SPY with iSVM returns the best results in all metrics with the use of $X$ feature set instead of $Z$ adjustment set. The use of $X$ extremely outperforms in any metric, with more significant gains in the almost elimination of any leakage of treated in the 'reliable controls' and the increase of recall. Even SPY method alone using $X$ outperforms the combined solution of SPY and iSVM when they are trained on $Z$. In the sowing and fertilization datasets, recall and precision follow almost the same fluctuations as in the simulated data. However, leakage of treated units presents a slight increase with the use of the feature set $X$ in comparison with the adjustment set $Z$. Furthermore, the fertilization dataset, when SPY and iSVM are trained on $X$ presents very high treated leakage, with $8$ out of $11$ hidden treated units classified as controls, which risks significantly biasing the results of the following effect estimation task. A likely explanation is the very small sample size of spies and hidden treated units for the machine learning task.

In all datasets, the number of retrieved 'reliable controls' tend to be more if the larger $X$ feature set is used for PU learning - column \textit{'\# Selected'} in Table \ref{tab:pu-evaluation-report}. After the trimming of both groups' units based on propensity scores to secure treated and control units have overlapped propensities, it appears (in the column \textit{'(treated, control) '} of Table \ref{tab:pu-ate-before-after}) that bigger overlap between groups emerges in the more complex 2-steps setup (SPY plus iSVM with the use of $X$ feature set).
This is also visualized in Sec.~\ref{sec:pscores} of the Appendix , where the propensity scores overlap for all datasets are depicted.
% This is also visualized in Figure \ref{fig:pu-ps_after_pu}, where the propensity scores overlap in the four different methodological setups for the non-linear dataset are depicted. The Sec.~\ref{sec:pscores} of the Appendix contains propensity scores overlap for all datasets. 
Due to the skewed propensity scores distribution of inferred reliable controls even for the best case (i.s. SPY plus iSVM with $X$) of sowing and fertilization datasets, we apply asymmetric trimming to retain units with estimated propensity scores in the range $[0.1, 0.6]$. This decision ensures that we operate within a region of the covariate space where there is reasonable overlap between positively labeled treated units and inferred control candidates. However, this approach limits our estimand to an $ATE$ within this low- to moderate-propensity subpopulation. Consequently, our $ATE$ estimates are not directly generalizable to the full population or to high-propensity treated units, which are trimmed out due to lack of comparable control analogs. While this trimming reduces the risks of extrapolation and positivity violation, it introduces a form of selection bias and may underestimate the true heterogeneity of treatment effects across the broader covariate space.
So, in terms of proper classification/reliable controls identification, the full 2-step PU learning method outperforms the use of SPY alone in most cases and the superiority of use of $X$ feature set instead of $Z$ adjustment set for the PU learning is profound. In the Sec.~\ref{sec:explain} of the Appendix we included a slope chart per dataset that compares the top-15 features in terms of influence and their change between the iSVM trained on $Z$ and $X$, using model coefficients as feature importance.

However, as a final evaluation step given in reality we have control units for each dataset, we leverage various methods from different methodological backgrounds to estimate the causal $ATE$ of each treatment on the outcome of interest before and after the retrieval of control units through our proposed solution. Observing the results in Table~\ref{tab:pu-ate-before-after}, we easily summarize for the two simulated and the sowing datasets that the setup with SPY and iSVM model trained on the feature set $X$ succeeds in retrieving a statistically significant $ATE$ very close to that is retrieved by the confirmed treated \& control groups with all different causal estimators. Also, we observe that in the cases where $Z$ is used in PU learning, we have an underestimate of $ATE$ and lacking of statistical significance, probably due to extremely large trimming $[0.1, 0.3]$ that isolate our estimation to a limited subpopulation and also leads to a very small number of units left in both groups.
\vspace{-0.1em}  % reduce space after the table

For the fertilization dataset, the results are less stable (Table \ref{tab:pu-ate-before-after}). Using the real control units, we observe a positive ATE, though not statistically significant. After applying the PU learning method — specifically the more robust SPY plus iSVM plus $X$ scenario — we observe a slight increase in ATE estimates, with linear regression and IPW yielding statistically significant results ($p$-value < 0.05), and matching and T-Learner producing similar outcomes. However, these results should be interpreted with caution due to high treated leakage in this scenario.
Additionally, we applied our method to a real unlabeled dataset, enriched with available confirmed controls, to identify reliable units and estimate the ATE using only these controls (ignoring their actual treatment status) and the treated group. This dataset includes $616$ unlabeled units and $99$ confirmed controls. Using the SPY plus iSVM plus $X$ method and trimming the propensity scores to the range $[0.05, 0.6]$ to ensure overlap, we identified $67$ reliable controls, $55$ of which are confirmed controls.
Once again, the results align with those obtained using only real controls or engineered pseudo-unlabeled data (Table \ref{tab:pu-ate-before-after}). Specifically, for the real unlabeled dataset: linear regression estimated an $ATE$ of $1.641$ ($CI: [-0.393, 3.675], p = 0.112$); IPW estimated an $ATE$ of $1.255$ ($CI: [-0.613, 3.002], p = 0.113$); matching yielded an $ATE$ of $1.207$ ($CI: [-0.709, 2.528], p = 0.133$); and the T-Learner produced an $ATE$ of $0.877$ ($CI: [0.075, 1.679]$).
Overall, even in this less reliable setting of the fertilization dataset, the results suggest that our PU learning approach can effectively recover the true $ATE$ in the absence of a known control group.
\vspace{-1em}  % reduce space after the table

\section{Conclusions}
In this work, we proposed, implemented, and evaluated the use of PU learning as a reliable approach for identifying control units from unlabeled data, addressing the challenge of missing confirmed controls. Our experiments demonstrate that, under appropriate configurations—such as avoiding restrictions on effect estimation covariates when training the PU learner—this approach can be effective. Further work should examine the trimming–extrapolation trade-off, explore using $P(T=1|X)$ to enrich the treated group, apply more advanced PU methods, and test on causal effect benchmarks. Ultimately, we provide a simple, practical, and realistic solution that can unlock a wide range of quasi-experiments in earth, environmental, and agricultural sciences, especially given the growing availability of large-scale earth observation data.

% \begin{verbatim}
\begin{acks}
% \section{Acknowledgements}
We express our gratitude to Alexis Apostolakis for his informative presentation on PU learning during one of our journal club gatherings, which served as the initial spark for the conception of this idea. We also thank Selected Biogas Farsala SA for their collaboration and provision of data for one of the use cases. This work was supported by the project “Climaca” (ID: 16196), carried out within the framework of the National Recovery and Resilience Plan Greece 2.0, funded by the European Union – NextGenerationEU.
\end{acks}

\bibliographystyle{ACM-Reference-Format}
\bibliography{main}

% %% If your work has an appendix, this is the place to put it.
% \appendix
\clearpage

\begin{appendix}

% \usepackage{booktabs}
% \usepackage{graphicx}
\begin{table*}[!hb]
\centering
\resizebox{\textwidth}{!}{%
\begin{tabular}{@{}llll|llll@{}}
\toprule
\multicolumn{4}{l|}{\textbf{Sowing}} &
  \multicolumn{4}{l}{\textbf{Fertilization}} \\ \midrule
\textbf{Feature(s) name} &
  \textbf{Variable/Vertex} &
  \textbf{Source} &
  \textbf{Set} &
  \textbf{Feature(s) name} &
  \textbf{Variable/Vertex} &
  \textbf{Source} &
  \textbf{Set} \\ \midrule
LOW, HIGH &
  Weather on sowing date &
  Nearest weather station &
  Z &
  SOC\_prediction &
  Topsoil organic carbon (2022) &
  NOA ML model &
  Z \\
peak\_ndvi, trapezoidal\_ndvi\_sow2harvest &
  Crop growth &
  NDVI via Sentinel-2 &
  X &
  accum\_precip\_total\_m &
  \begin{tabular}[c]{@{}l@{}}Accumulated precipitation\\ (2022-09-01 - 2023-05-01)\end{tabular} &
  ERA5-land &
  Z \\
ndwi\_sowingday &
  Soil moisture on sowing date &
  NDWI via Sentinel-2 &
  Z &
  gdd\_base\_0 &
  \begin{tabular}[c]{@{}l@{}}Growing degree days\\ (2022-09-01 - 2023-05-01)\end{tabular} &
  ERA5-land &
  Z \\
clay\_mean, silt\_mean, sand\_mean &
  Topsoil properties &
  Map by ESDAC &
  Z &
  ndmi\_trapezoidal\_area &
  \begin{tabular}[c]{@{}l@{}}Accumulated vegetation moisture content\\ (2022-12-01 - 2023-05-01)\end{tabular} &
  NDMI via Sentinel-2 &
  Z \\
occont\_mean &
  Topsoil organic carbon &
  Map by ESDAC &
  Z &
  sum\_soil\_moisture\_mean &
  \begin{tabular}[c]{@{}l@{}}Soil moisture\\ (2022-09-01 - 2023-05-01)\end{tabular} &
  ERA5-land &
  Z \\
var\_\{ST\_402, ..., ELPIDA\} &
  Seed variety &
  Farmers' Cooperative &
  Z &
  st\_\{Cambisols, ..., st\_Luvisols\} &
  Soil type &
  SoilGrids World Reference Base(2006) Soil Groups &
  Z \\
ratio &
  Geometry of field &
  Farmers' Cooperative &
  Z &
  POI\_\{7807, ..., 10486\} &
  Seed variety &
  LPIS from NPA &
  Z \\
lat, lon, perimeter, field\_area &
  Location \& area of field &
  Farmers' Cooperative &
  X &
  eomi\_\{1-4\}, nbr2 - stats: max, min, mean, std, skew &
  Exogenous organic matter indexes &
  via Sentinel-2 &
  X \\
hday\_sin, hday\_cos, len\_season &
  Harvest date \& cultivation period &
  Farmers' Cooperative &
  X &
   &
   &
   &
   \\ \midrule
prediction &
  \textbf{\begin{tabular}[c]{@{}l@{}}Treatment\\ (sown on recommended or not date)\end{tabular}} &
  Farmers' Cooperative, RS &
  T &
  TREATMENT &
  \textbf{\begin{tabular}[c]{@{}l@{}}Treatment\\ (apply or not fertilizer)\end{tabular}} &
  Fertilizer company &
  T \\
yield21 &
  \textbf{\begin{tabular}[c]{@{}l@{}}Outcome \\ (Yield)\end{tabular}} &
  Farmers' Cooperative &
  X &
  ndvi\_trapezoidal\_area &
  \textbf{\begin{tabular}[c]{@{}l@{}}Outcome\\ (produced biomass proxy for 2023-05-01 - 2023-06-30)\end{tabular}} &
  NDVI via Sentinel-2 &
  X \\ \bottomrule
\end{tabular}%
}
\caption{Variables used for causal modeling in the sowing and fertilization case studies, categorized by their role (Z: adjustment set, X: features set, T: treatment, Y: outcome) and source.}
\label{tab:rworld-data-vars}
\end{table*}

\section{Real-World Datasets}
\label{sec:rworld-data}

For the sowing case, we follow the causal graph construction presented in Tsoumas et al. \cite{tsoumas2023evaluating}, which guides the selection of relevant variables. In the fertilization case, we model the effect of fertilizer application on biomass production during the final phenological stages, incorporating both decisional and biological confounders. These confounders are selected based on a merged causal graph that combines the canonical fertilization–yield graph from Pearl’s book \cite{pearl2018book} with a more detailed soil-process causal graph from \cite{sharma2023knowledge}. Additionally, exogenous organic matter indexes, included as $X$ variables, are retrieved from the literature as features indicative of fertilization activity detectable by Sentinel-2 \cite{dodin2023sentinel, kalogeras2025monitoring}. As detailed in Table~\ref{tab:rworld-data-vars}, adjustment variables ($Z$) cover the cultivation period up to 01/05/2023, the treatment variable ($T$) indicates whether fertilizer was applied at least once between 01/09/2022 and 01/05/2023, and the outcome ($Y$) corresponds to NDVI-derived trapezoidal area between 01/05/2023 and 30/06/2023, capturing the produced biomass at the end of the season.

\section{Simulation Datasets}
\label{sec:sim-setup}

We simulate data under two different structural assumptions. A \textbf{linear} setup and a \textbf{nonlinear} setup. In both cases, we generate $n = 1000$ samples.

\subsection*{Observed, Unobserved and Latent Variables}
\begin{align*}
    X_1, X_2, X_{11}, X_{22}, X_3, X_4, X_8 &\sim \mathcal{N}(0, 1) \\
    U_1, U_2, U_3 &\sim \mathcal{N}(0, 1) \\
    X_5 &= 0.6 U_1 + 0.6 U_2 + \varepsilon_5,\quad \varepsilon_5 \sim \mathcal{N}(0, 0.1^2)
\end{align*}

\subsection*{Linear Setup}

\paragraph{\textbf{Treatment Assignment}}
\begin{align*}
    \text{logit}\left(P(T = 1)\right) &= 1.2 X_1 + 0.8 X_2 + 1.4 X_{11} + 0.6 X_{22} + 0.7 X_4 + 1.0 U_1 + 1.0 U_3 \\
    T &\sim \text{Bernoulli}\left(\sigma(\cdot)\right),\quad \sigma(z) = \frac{1}{1 + e^{-z}}
\end{align*}

\paragraph{\textbf{Mediator}}
\begin{align*}
    M &= 1.5 T + 0.7 X_8 + 0.5 X_2 + 0.3 X_{22} + \varepsilon_M,\quad \varepsilon_M \sim \mathcal{N}(0, 0.1^2)
\end{align*}

\paragraph{\textbf{Outcome}}
\begin{align*}
    Y &= 2.0 M + 0.7 X_1 + 0.5 X_{11} + 0.6 X_3 + 1.0 U_2 + 1.0 U_3 + \varepsilon_Y,\quad \varepsilon_Y \sim \mathcal{N}(0, 0.1^2)
\end{align*}

\paragraph{\textbf{Proxy Variables}}
\begin{align*}
    X_7 &= 1.2 M + \varepsilon_{X_7},\quad \varepsilon_{X_7} \sim \mathcal{N}(0, 0.1^2) \\
    X_9 &= 1.0 T + 1.0 Y + \varepsilon_{X_9},\quad \varepsilon_{X_9} \sim \mathcal{N}(0, 0.1^2)
\end{align*}

\subsection*{Nonlinear Setup}

\paragraph{\textbf{Treatment Assignment}}
\begin{align*}
    \text{logit}\left(P(T = 1)\right) &= 1.2 \tanh(X_1) + 0.8 \sin(X_2) + 1.4 \tanh(X_{11}) + 0.6 \sin(X_{22}) \\
    &\quad + 0.7 \tanh(X_4) + 1.0 U_1 + 0.8 U_3 X_1 \\
    T &\sim \text{Bernoulli}\left(\sigma(\cdot)\right)
\end{align*}

\paragraph{\textbf{Mediator}}
\begin{align*}
    M &= 1.5 T + 0.7 \sqrt{|X_8|} + 0.5 \log(1 + |X_2|) + 0.3 X_{22} + 0.1 X_2 X_8 + \varepsilon_M
\end{align*}

\paragraph{\textbf{Outcome}}
\begin{align*}
    Y &= 2.0 M^2 + 0.7 X_1 + 0.5 X_{11} + 0.6 \sin(X_3) + 0.2 X_3^2 + 1.0 U_2 + 1.0 U_3 + \varepsilon_Y
\end{align*}

\paragraph{\textbf{Proxy Variables}}
\begin{align*}
    X_7 &= 1.2 M + \varepsilon_{X_7} \\
    X_9 &= 1.0 T + 1.0 Y + \varepsilon_{X_9}
\end{align*}

In both setups, the variables $X_7$ and $X_9$ serve as proxies. $X_7$ for the mediator $M$, and $X_9$ as a collider involving both $T$ and $Y$.

\section{Propensity scores}
\label{sec:pscores}

Following the figures for each experimental dataset with an overview of how propensity score overlap varies in the current experimental dataset when using different predictor sets (i.e $X,Z$) and methods (i.e. SPY, SPY + iSVM). This highlights how the choice of predictor variables and estimation approach affects the positivity assumption and common support.

\begin{figure*}[]
  \centering
  \includegraphics[width=\textwidth]{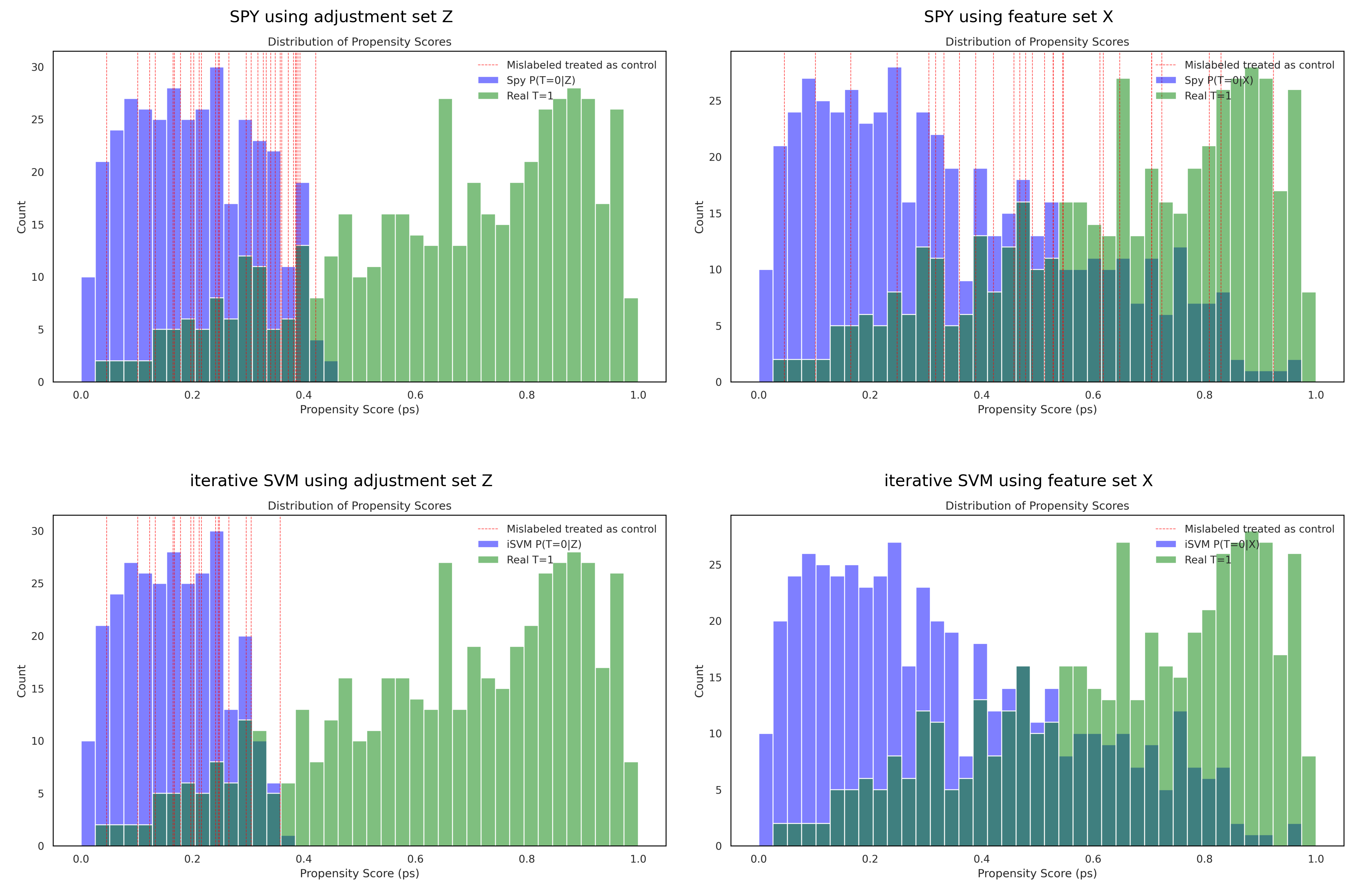}
  \caption{Propensity scores of 4 different combinations on linear experimental dataset}
  % \label{fig:pu-framework}
\end{figure*}

\begin{figure*}[]
  \centering
  \includegraphics[width=\textwidth]{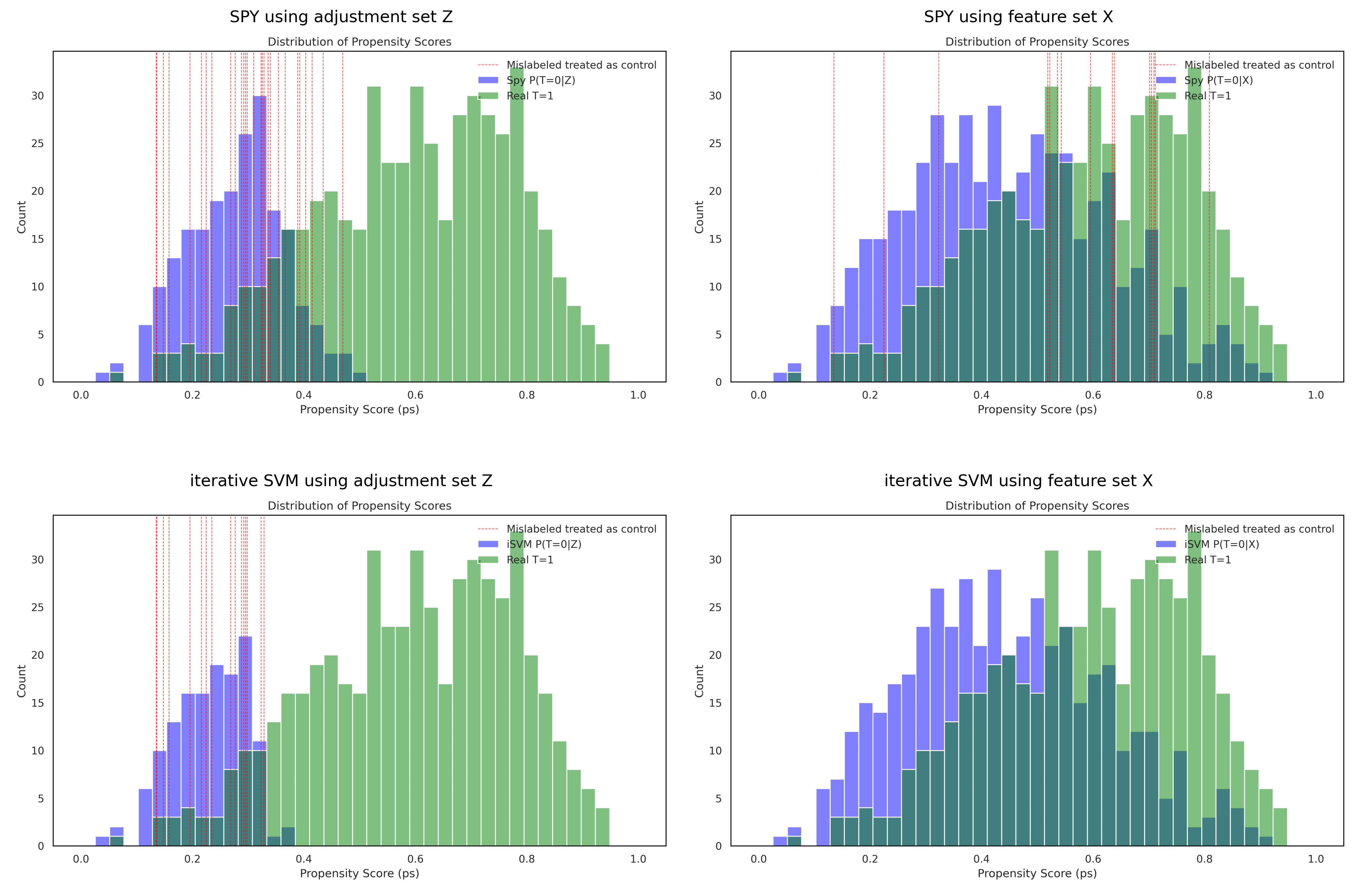}
  \caption{Propensity scores of 4 different combinations on non-linear experimental dataset}
  % \label{fig:pu-framework}
\end{figure*}

\begin{figure*}[]
  \centering
  \includegraphics[width=\textwidth]{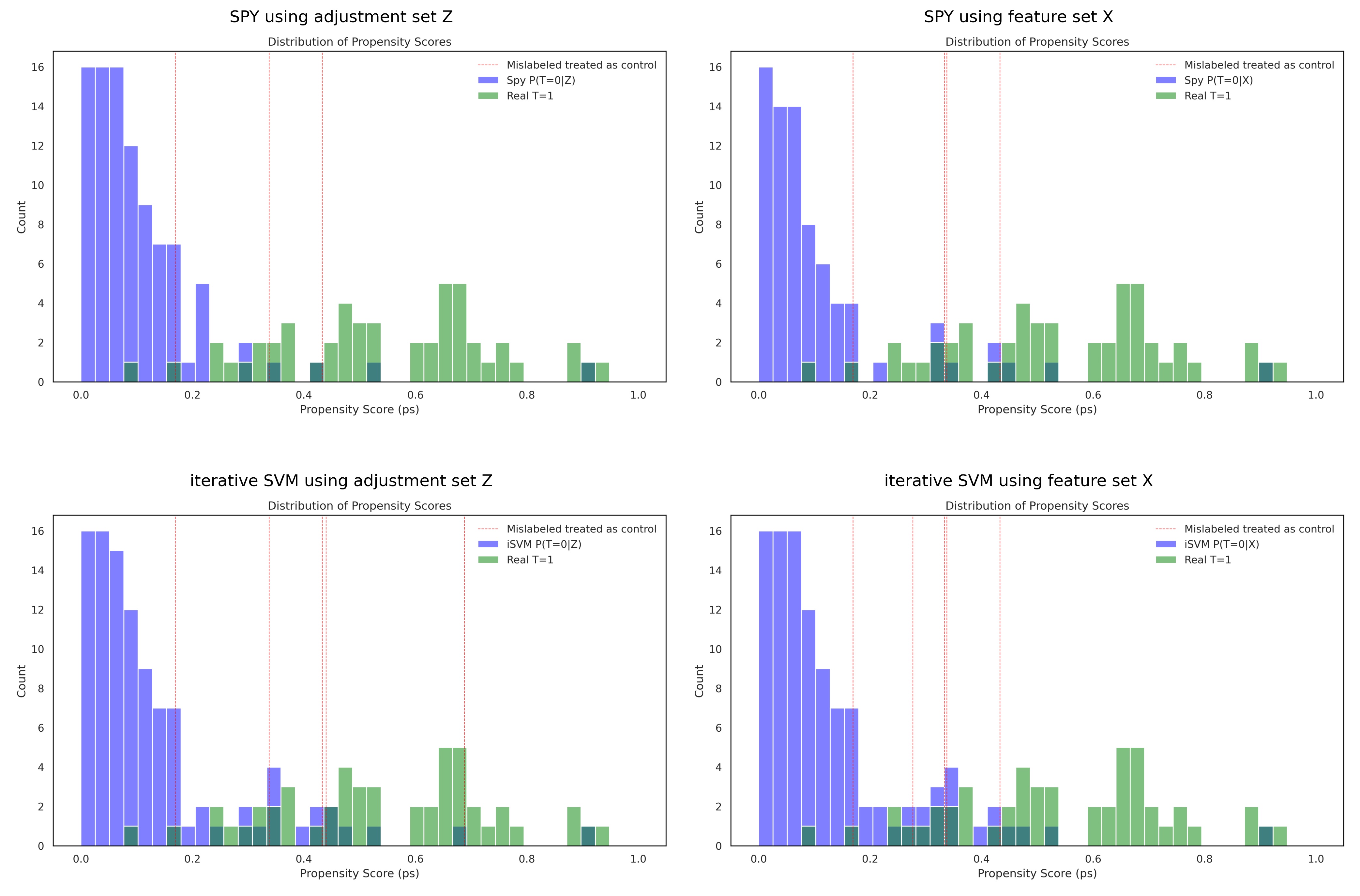}
  \caption{Propensity scores of 4 different combinations on  experimental dataset regarding optimal sowing}
  % \label{fig:pu-framework}
\end{figure*}

\begin{figure*}[]
  \centering
  \includegraphics[width=\textwidth]{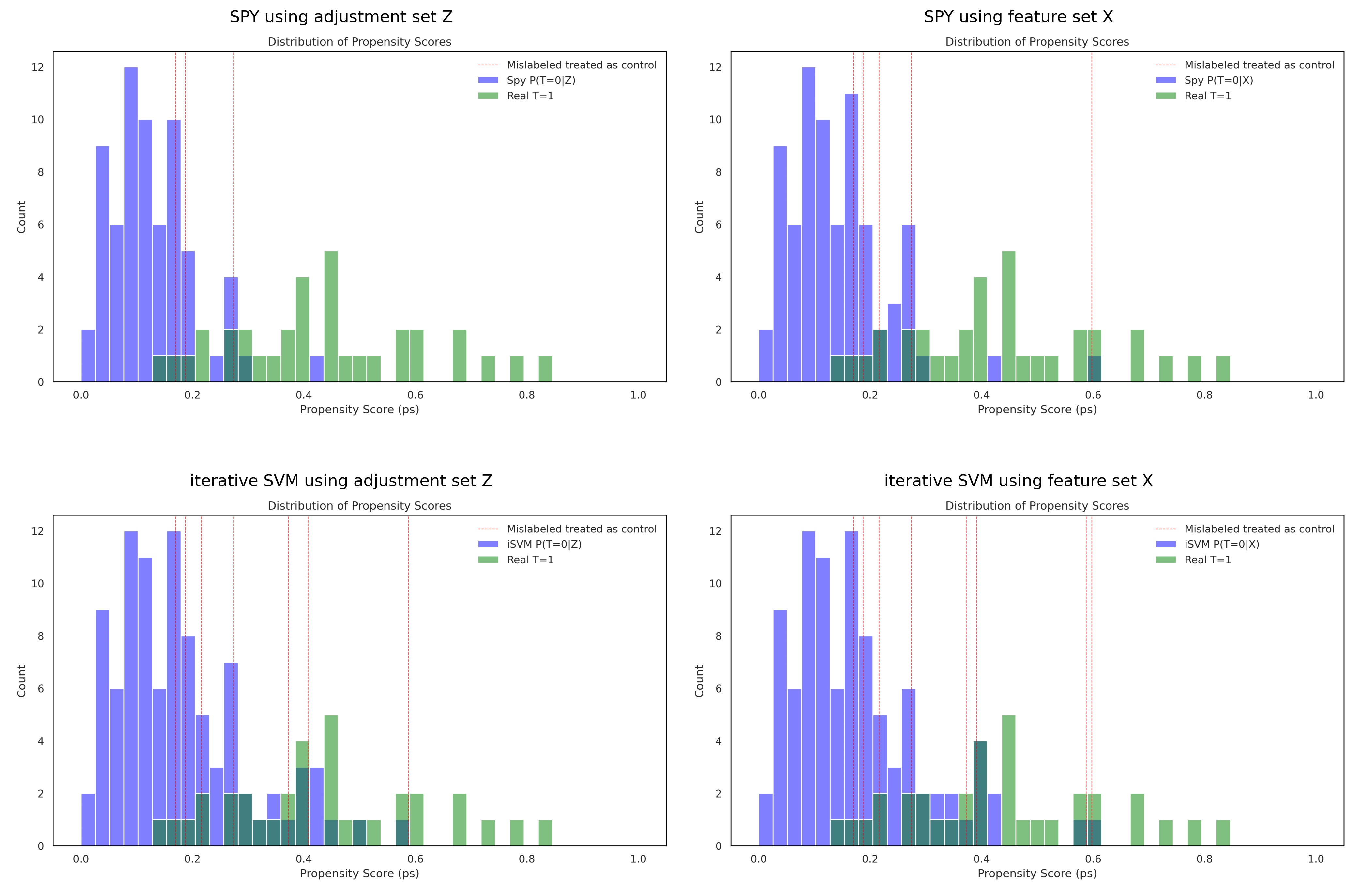}
  \caption{Propensity scores of 4 different combinations on  experimental dataset regarding digestate fertilization application}
  % \label{fig:pu-framework}
\end{figure*}
\clearpage
\section{Interpretability and Model Comparison}
\label{sec:explain}
To investigate and compare the internal decision mechanisms of two Support Vector Machine (SVM) classifiers, we examine the learned feature coefficients from each model. Both classifiers share a common architecture but differ in the feature sets used for training. We leverage the linear nature of the SVM (with a linear kernel) to extract and interpret feature importance directly from the model coefficients. These coefficients represent the influence of each feature on the decision boundary—positive values contribute toward predicting the positive class (e.g., sowing on a good day), and negative values contribute toward the negative class.

To visually communicate how feature importance shifts between the two models, we use a slope chart. Each line in the chart corresponds to a feature, connecting its normalized coefficient in the model trained on adjustment set $Z$ (left axis) to its corresponding value in the model trained of the superset of it, the feature set $X$ (right axis). Features that are newly introduced in the superset $X$ appear at zero on the left and are highlighted to emphasize their emergence in the decision function. Such comparative interpretability enhances our understanding of how additional information reshapes the model’s decision boundary..

\begin{figure*}[]
  \centering
  \includegraphics[width=\textwidth]{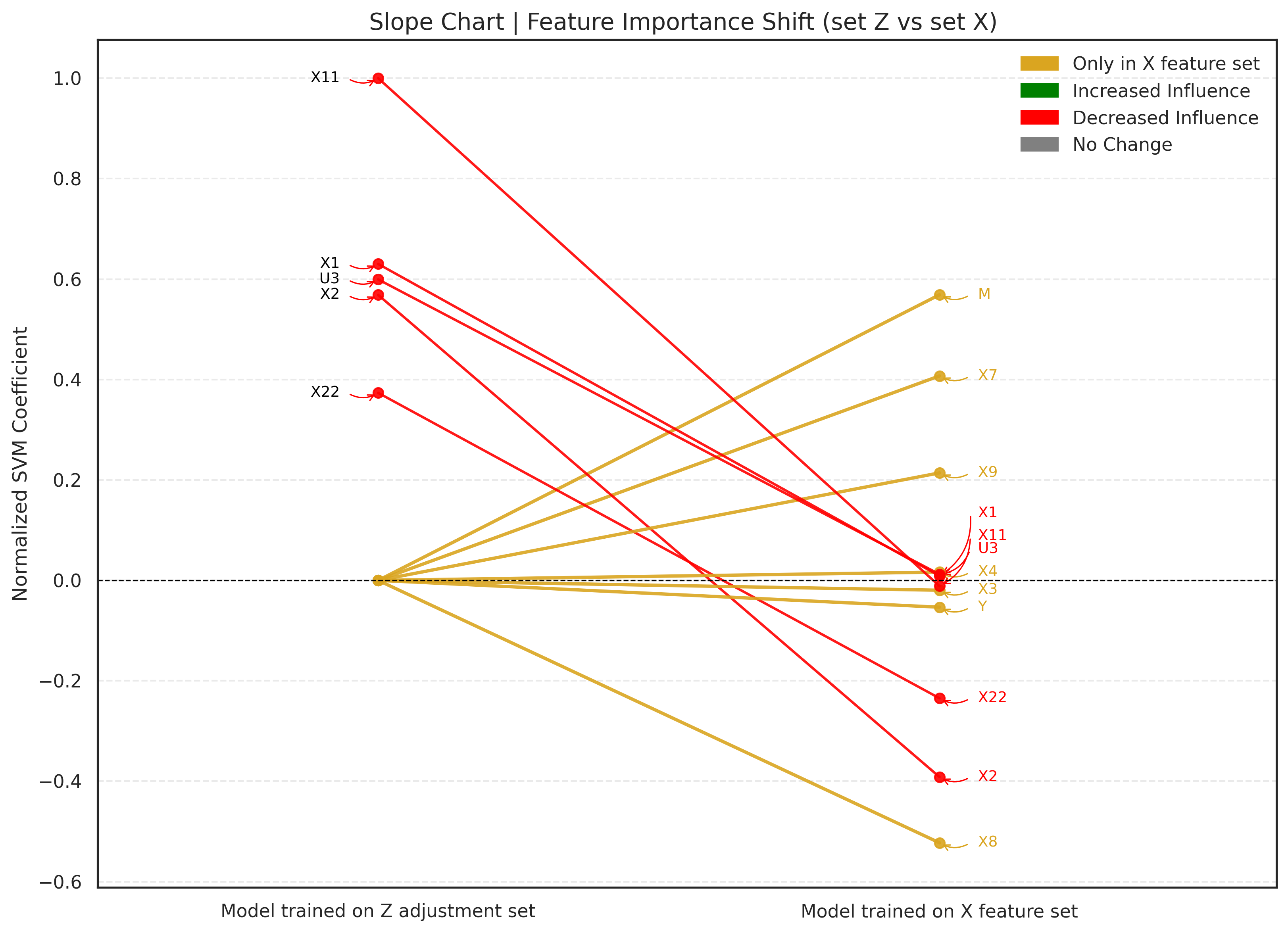}
  \caption{Interpretability and Model Comparison via SVM Coefficient Slope Chart for linear dataset}
  % \label{fig:pu-framework}
\end{figure*}

\begin{figure*}[]
  \centering
  \includegraphics[width=\textwidth]{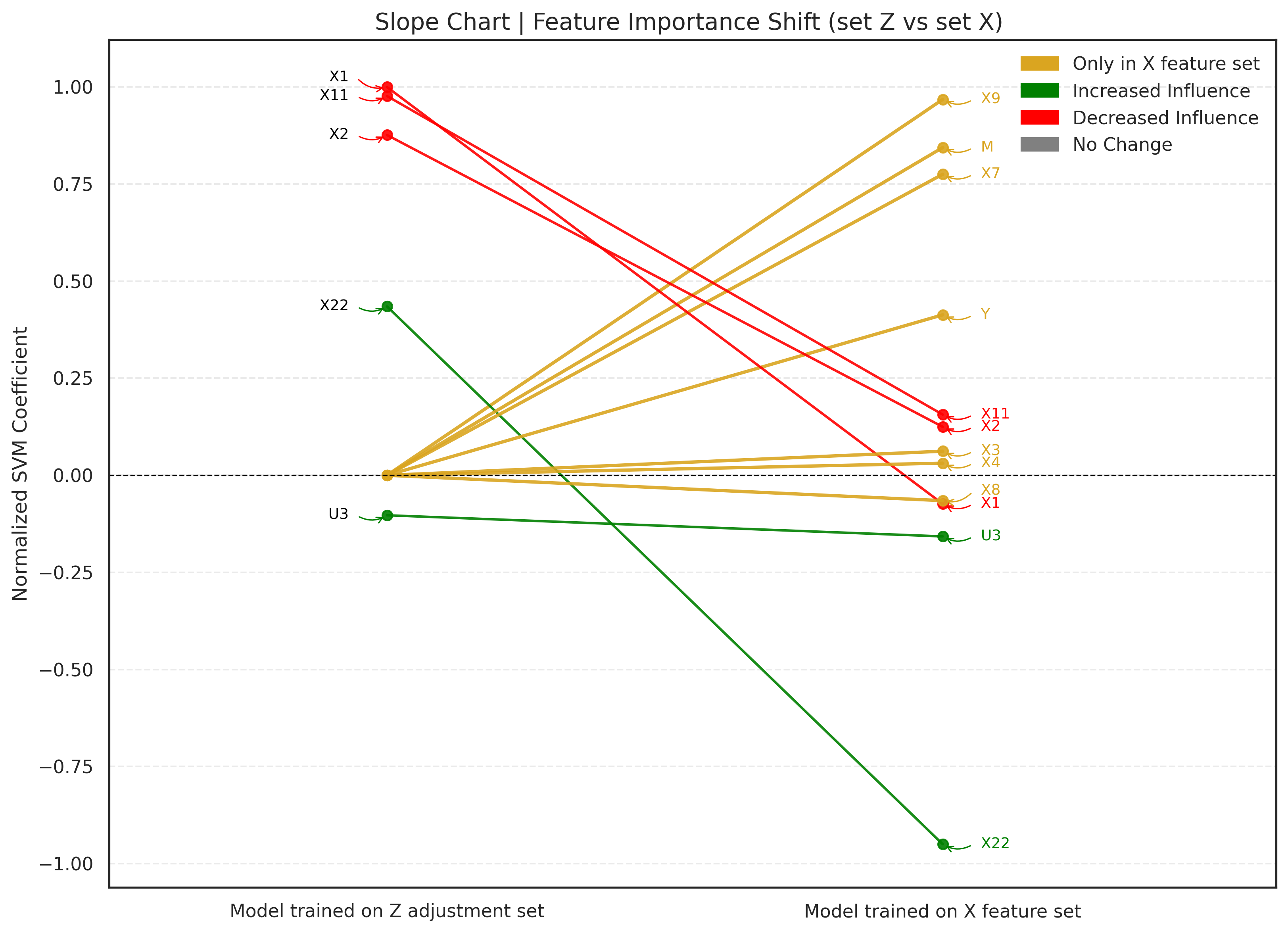}
  \caption{Interpretability and Model Comparison via SVM Coefficient Slope Chart for non-linear dataset}
  % \label{fig:pu-framework}
\end{figure*}

\begin{figure*}[]
  \centering
  \includegraphics[width=\textwidth]{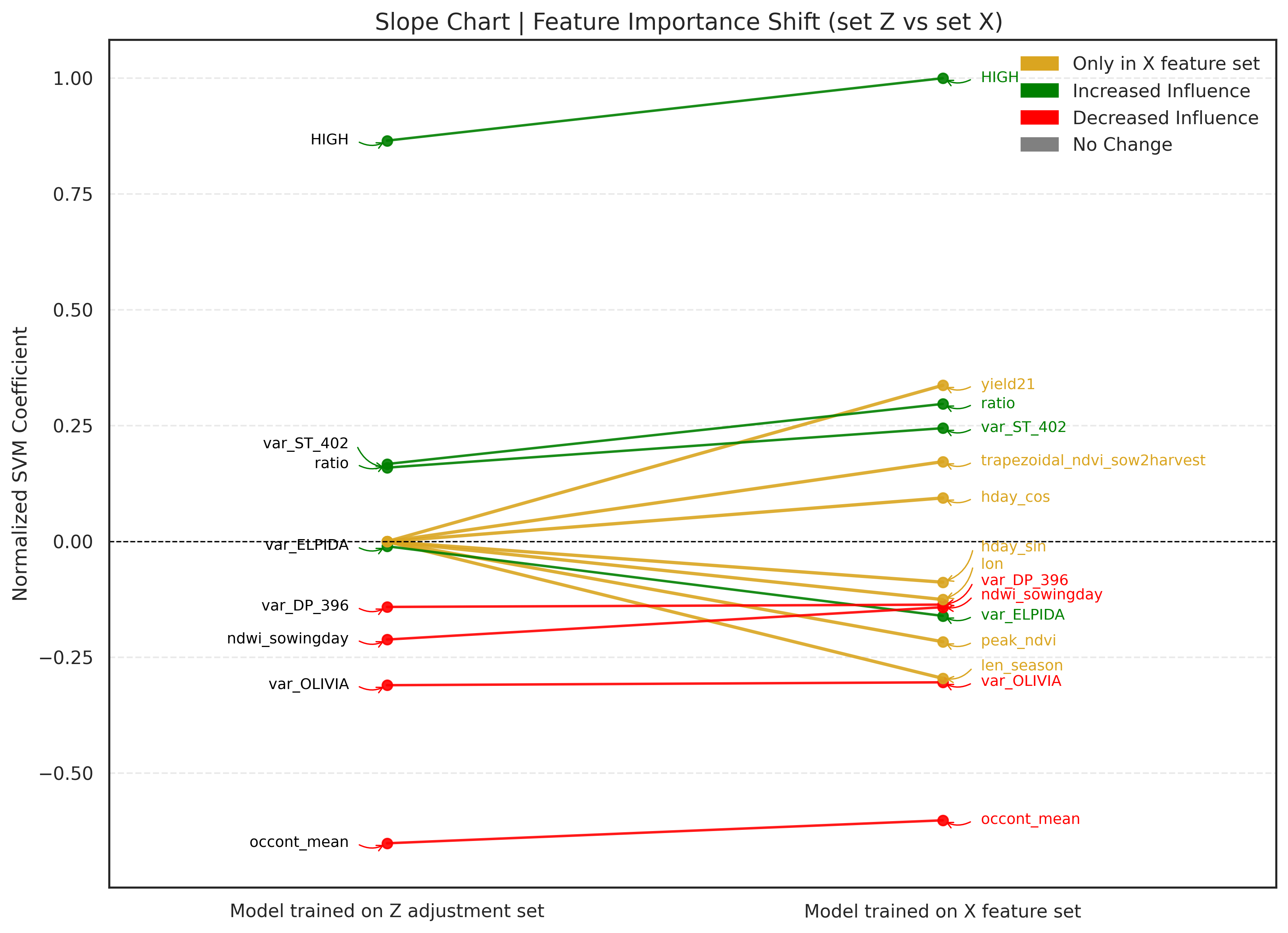}
  \caption{Interpretability and Model Comparison via SVM Coefficient Slope Chart for dataset regarding optimal sowing}
  % \label{fig:pu-framework}
\end{figure*}

\begin{figure*}[]
  \centering
  \includegraphics[width=\textwidth]{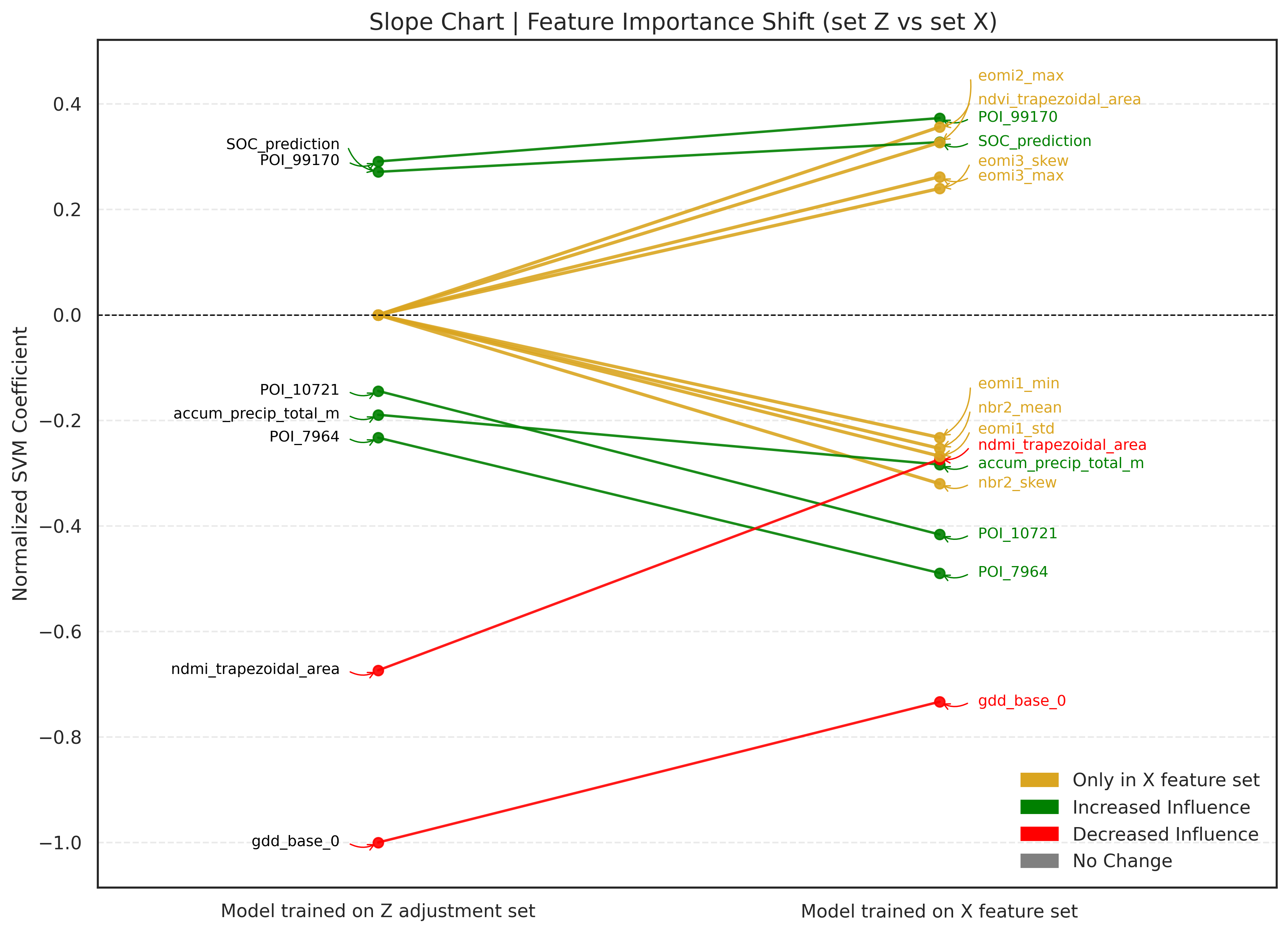}
  \caption{Interpretability and Model Comparison via SVM Coefficient Slope Chart for dataset regarding digestate fertilization application}
  % \label{fig:pu-framework}
\end{figure*}








\end{appendix}
\end{document}